\documentclass[letterpaper]{article} 
\usepackage{aaai2026}  
\usepackage{times}  
\usepackage{helvet}  
\usepackage{courier}  
\usepackage[hyphens]{url}  
\usepackage{graphicx} 
\usepackage{natbib}  
\usepackage{caption} 
\usepackage{algorithm}
\usepackage{algorithmic}

\usepackage{newfloat}
\usepackage{listings}
\DeclareCaptionStyle{ruled}{labelfont=normalfont,labelsep=colon,strut=off} 
\floatstyle{ruled}
\newfloat{listing}{tb}{lst}{}
\floatname{listing}{Listing}
\usepackage{algorithm}
\usepackage{algorithmic}

\usepackage{xcolor}

\usepackage{newfloat}
\usepackage{listings}
\floatstyle{ruled}
\newfloat{listing}{tb}{lst}{}
\floatname{listing}{Listing}
\title{Structure of Basic Human Values in Russian Social Media}
\author{
    Maria Milkova\textsuperscript{\rm 1},
    Maxim Rudnev\textsuperscript{\rm 2}
}

\affiliations{
    \textsuperscript{\rm 1}Independent researcher, Lisbon, Portugal\\
    \textsuperscript{\rm 2}UCL Institute of Education, London, United Kingdom\\
}

\begin{document}

\maketitle

\begin{abstract}
What is known about the values of national populations rests almost entirely on questionnaires, which prompt respondents to rate researcher-supplied descriptions of each motivation. Social media instead records values as they are invoked spontaneously, in communication addressed to a real audience. We measure value expression in broadly sampled, heterogeneous Russian-language social-media data, where value-relevant content is sparse and often ambiguous, and read the result against the survey-based profile of the same population. To address subjectivity in value interpretation, we compare alternative LLM-prompt configurations against expert judgments and use error-driven calibration to select a theory-based operationalization. Using VKontakte data collected from one million randomly generated user IDs, comprising 7.5 million public text posts, we develop a multi-stage framework combining candidate selection, repeated LLM annotation, and transfer to an XLM-RoBERTa encoder for large-scale prediction (a macro PR-AUC of .60 on the expert benchmark, with continuous scores positively correlated with graded expert confidence across all ten values). The circular structure proposed by Schwartz is largely recovered (congruence .95 at the post level, .94 at the user level). The aggregate profile agrees with survey findings at the anchors --- Benevolence highest and broadly distributed across users, Power lowest --- but shows relatively elevated Hedonism and relatively low Universalism. Universalism is further displaced structurally, co-expressed with Security and Tradition and concentrated in the output of highly active users.
\end{abstract}


\section{Introduction}

Studying value expression in social media provides a common theoretical lens for characterizing what is important and desirable in a given population \citep{shifman2026expression}. What is currently known about the values of national populations, including their relative importance and the circular motivational structure that organizes them \citep{schwartz1992universals,schwartz2012refining}, rests almost entirely on a single measurement mode. The Schwartz Value Survey, the Portrait Values Questionnaire, and the Human Values Module all \emph{prompt} the respondent: the researcher supplies verbal stimuli of each motivational type, implicitly rendering all ten equally salient, and requires a response to each \citep{schwartz2001extending,davidov2008bringing}. Values expressed on social media, by contrast, are unprompted, incidental to ordinary communicative goals, addressed to a real audience rather than an investigator, and surfacing only when something in the situation makes them relevant. The first mode records values as \emph{prompted} \citep{lake2026values}; the second, values as spontaneously \emph{expressed}. Convergence between them is not guaranteed. Prompted methods are affected by limited introspective access \citep{nisbett1977telling}, socially desirable responding \citep{paulhus2002socially}, and the reference-group effect \citep{heine2002whats}.  Measures based on spontaneous expression are subject to the unequal lexical availability of different motivations and to communication norms \citep{ponizovskiy2020development}. These differences can contribute to the discrepancy between value profiles, as the two measurement modes tap into different aspects of value motivation.

This makes spontaneous expression substantively interesting rather than merely convenient. Whether the circular structure re-emerges in unprompted discourse bears on whether it resides in the organization of motivation itself or partly in the architecture of the instrument. And because digital traces afford population-scale, non-reactive, continuous observation \citep{kosinski2013private}, they are especially attractive where survey response bias and self-censorship are documented concerns \citep{kuran1995private,frye2017putin} --- as in Russia, whose survey-based value profile is well characterized \citep{magun2012basic} and therefore available as a point of comparison. Our aim is accordingly substantive: to describe aggregate value expression and value co-expression in Russian-language social media at population scale. Realizing that aim, however, requires measurement that does not yet exist for this kind of data, and a large part of this paper is therefore methodological, in service of the substantive question.

Recent work has advanced the automatic recognition of human values in text, including through the use of LLMs for social-media content \citep{epstein2026whose,starovolsky2025value}. Existing computational studies of value expression have typically worked within bounded textual settings that are filled with value expressions, such as ideological arguments, political manifestos, or news \citep{kiesel2023valueeval,mirzakhmedova2024touche23,scharfbillig2026valuesml,borenstein2025investigating}. More generally, what can be inferred from social-media data depends on how the underlying data are selected \citep{tufekci2014big,vicente2023sampling}. User-level sampling of public self-expression provides a less topically constrained view of social media activity, but consequently yields a highly heterogeneous corpus in which value-relevant content is dispersed across many different forms of communication. Measuring value expression in such heterogeneous data presents several challenges. First, value-relevant content is sparse. Second, the interpretation of value expression, already shown to be potentially ambiguous \citep{epstein2026whose,falk2025mining}, becomes particularly difficult in this setting, as users’ public self-expression may range from greetings, poetry, and quotations to news reposts and political statements.
Operationalizing values across such diverse contexts requires conceptual clarity and methodological transparency, particularly when obtaining annotations for training supervised models. Third, value attribution is often
graded, calling for continuous measures that retain variation in the certainty of attribution.

We address these challenges by developing a multi-stage framework for measuring value expression in large-scale Russian-language social-media data, grounded in Schwartz's theory of basic human values \citep{schwartz1992universals}. We compare alternative LLM-prompt configurations and calibrate the annotation procedure against expert judgments, with the aim of selecting a theory-based operationalization. Starting from public timelines collected through one million randomly generated user IDs, the framework includes: (i) selection of value-related content, (ii) repeated LLM annotation under the selected configuration, aggregated into soft labels that retain graded variation in value attribution, and (iii) transfer of these annotations to an encoder model that produces value-expression scores. We use data from VKontakte (VK), a prominent Russian social-media platform, spanning 2007–2025 and providing a large-scale record of user self-expression.

Substantively, we characterize the \emph{value climate} of public self-expression in Russian social media: which motivational concerns are voiced, how often, and which are voiced together.  Although we did not expect a close approximation of the Schwartz theory in the social media data as it is a spontaneous expression filtered through the communication norms and biased towards more frequent posters, we expected a moderate congruence at the posts level and high congruence at the individual level. Methodologically, our contribution is to extend corpus-level measurement of value expression to broadly sampled public user timelines, explicitly addressing interpretive variation in value operationalization. Taken together, the two allow the resulting picture of spontaneous expression to be read against what the survey tradition reports for the same population, and to interpret the differences.

\section{Related work}

Human values have been studied in social media both as latent characteristics of users and as properties expressed in online content. Earlier studies treated digital traces as indicators of individual value orientations \citep{sun2014post,mukta2019temporal}. More recent work has focused increasingly on values expressed in communication itself. Such work spans diverse communicative settings, including lifestyle-oriented
posts \citep{hallinan2023values,trillo2021freedom}, far-right memes \citep{trillo2021memetic}, influencer content on Instagram \citep{gonzalez2025values} and TikTok \citep{starovolsky2025value}, and online communities on Reddit \citep{borenstein2025investigating}.

From a computational perspective, identifying expressed values is typically formulated as a multi-label classification problem, in which texts are assigned one or more predefined value categories \citep{kiesel2023valueeval,scharfbillig2026valuesml}. At the same time, the attribution of value expression is inherently interpretive, and recent work has questioned the assumption of a single objective label for value expression \citep{falk2025mining}. \citet{epstein2026whose} address this variation by developing personalized classifiers calibrated to individual raters’ perceptions. These findings connect value detection to a
broader perspectivist literature on subjective annotation, which treats disagreement as potentially meaningful variation arising from ambiguity, context, and perspective \citep{plank2022problem,davani2022dealing,fleisig2024perspectivist}. Modeling value expression therefore requires attention not only to classification accuracy, but also to how alternative interpretations are
represented in the annotation and prediction process.

The increasing use of LLMs as annotators introduces an additional interpretive layer into this problem. LLM annotation outcomes vary across tasks and models \citep{ziems2024large,pangakis2025keeping} and are sensitive to prompt framing and perspective specification \citep{schafer2025demographics}. In multilingual settings, annotation outcomes may also vary with instruction language and cultural context \citep{cui2025bias}. Recent work therefore emphasizes explicit validation of LLM-generated annotations \citep{pangakis2025keeping,calderon2025alternative}.
Taken together, modeling human value expression at scale requires procedures that account for interpretive subjectivity, ground value attribution in theory, and use appropriate evaluation and validation criteria.

\section{Data and Methods}

\subsection{Data Collection}

The data were collected from VK social media using the VK API. One million user IDs were sampled at random, independently of users’ posting activity, topics, or community membership. Due to limitations of the VK API, data were retrieved in several batches between August 2022 and January 2025\footnote{VK assigns numeric user identifiers sequentially. At the start of the study, assigned user IDs extended to approximately 710 million. Sampling frame therefore comprised integers from 1 to 710,000,000. Data were retrieved in seven batches. Previously sampled IDs were re-queried in later collection rounds to capture newly published posts from profiles that remained public.}. Of sampled IDs, 155,629 (15.6\%) corresponded to profiles that were publicly accessible and contained visible text posts at the time of collection\footnote{The remaining IDs included private profiles, deleted or otherwise inaccessible accounts, and profiles with no textual content.}. For these users, we collected all available posts from their timelines, including original posts, reposts, and link titles. The resulting data span 2007-January 2025 and contain 7,498,657 posts. We retained texts containing Cyrillic characters and at least two words. After this filtering, the corpus comprised 5,561,547 posts from 122,668 users.

\subsection{Candidate post selection}

\subsubsection{Data filtering.} The VK data were noisy, so at the first step we filtered out spam posts, such as advertisements and auto-generated messages from VK applications. We also filtered out posts written on behalf of companies or organizations, including political. To do so, we constructed a rule-based filter using high-frequency lexical patterns and duplicated text fragments characteristic of spam. The resulting rule-based labels were then used as weak supervision to train a logistic-regression classifier based on TF-IDF unigram and bigram features, which identified additional spam and non-personal content not captured by the initial rules. After this stage, 3,583,536 posts from 103,242 users remained.

\subsubsection{Detecting value-expressive text posts.} Determination of whether a post is value-expressive or not is a necessary first step before the post can get its value profile. Value-expressive posts were detected in three steps, following the procedure proposed in our earlier work \citet{milkova2023detecting}. First, we annotated a smaller subset of data; second, we trained a model based on this data which is capable of predicting value-expressiveness; finally, we used this trained model to detect value-expressive posts in the full dataset. The annotation set consisted of 5,035 posts, each evaluated independently by three ChatGPT-3.5 runs and three crowd-worker judgments. The crowd annotations were provided by 304 paid Russian-speaking workers recruited through the Toloka platform\footnote{URL: https://toloka.ai.}, who received a predefined annotation guide (see Appendix~\ref{app:toloka} for details). A separate subset of held-out data (804 posts) was annotated by three experts and used to evaluate the annotation strategies. Relative to the expert majority, both ChatGPT and crowd-worker annotations achieved F1=.65, but exhibited different error patterns. ChatGPT provided higher recall, whereas crowd judgments were more conservative. The goal of this stage was to minimize the exclusion of potentially value-expressive posts. Therefore, we combined the two sources, using ChatGPT labels as the baseline and overriding positive labels when crowd workers unanimously classified a post as non-value-relevant. The resulting training set contained 34.4\% value-expressive posts. These labels were used to train an SVM classifier based on RuBERT-tiny embeddings. On held-out data, it achieved F1=.77 for the positive class.

\subsubsection{Detecting politically oriented posts.} Expert review of the preliminary value-relevance classifications indicated that politically oriented posts could contain value-relevant signals even when they were not selected by the general value-expression classifier (e.g. news reposts). To reduce the systematic exclusion of such content, we developed an additional classifier for politically oriented posts. We first randomly selected a dataset of 84,025 posts and compiled a list of politically relevant keywords. This list included names of political leaders, countries, regions, political organizations, and a range of terms related to military topics, including informal and slang words. Keyword matching identified 4,891 candidate posts, which were manually reviewed; 1,808 were confirmed as political, corresponding to 2.1\% of the sampled posts. These annotations were split into training and test sets using an 80/20 split. A linear SVM classifier with DistilRuBERT embeddings achieved an F1 score of .82 for the positive class.

After the data preprocessing and filtering described above, the final dataset of value- and/or politically-expressive posts consisted of 1,105,085 posts belonging to 53,606 users. 

\subsection{Value Annotation and Encoder Transfer}

The annotation and evaluation procedures, as well as the expert verification design, build on our previous methodological study \citet{milkova2026measuring}.

\subsubsection{Expert annotation.} To calibrate and validate the value-annotation procedure, we constructed two expert benchmarks based on posts retained after the candidate-selection stage. Posts were annotated for the ten basic values defined by Schwartz’s theory. The main (first) benchmark consisted of 1,000 posts independently annotated by three experts in a multi-label setting. Before the main annotation, the experts completed a pilot task of 80 posts and provided justifications for their decisions to establish a common interpretation of the value definitions. A second benchmark consisted of 2,000 posts annotated by a different group of three experts in a binary setting: for each value, 200 posts were randomly sampled and independently evaluated for the presence of that target value. This second dataset was used to assess the robustness of the final annotation configuration. All expert annotators were Russian-speaking researchers with expertise in human values, recruited through the authors' academic networks. They participated voluntarily and without compensation (see Appendix~\ref{app:expert_instruction} for the expert annotation instructions). Inter-annotator agreement was moderate in the multi-label benchmark (Fleiss’ $\kappa=.60$) and lower in the value-specific benchmark ($\kappa=.30$).
In line with recent work on subjective value annotation, we treat expert judgments as competing reasonable interpretations. For categorical evaluation, we use the expert majority label, for graded evaluation we retain the number of positive expert votes (0–3) as a measure of confidence that a specific value is expressed in the post. 

\subsubsection{LLM annotation.} We evaluated three LLM models, GPT-4, GPT-5.2, and Gemini-2.5-pro, and experimented with different prompts. We started with the baseline prompt containing short value definitions according to Schwartz’s theory. We then subsequently examined over- and under-attribution of values relative to expert judgments and used these error patterns to further calibrate the prompts. Definitions of over-attributed values were made more restrictive, whereas definitions of under-attributed values were expanded to better capture implicit expressions. The revised instructions also emphasized conceptual meaning over lexical cues, discouraged interpretations unsupported by the text, and clarified motivational relationships among values. We then evaluated a Russian-language version of the calibrated prompt to assess whether instruction language affected annotation behavior. The full prompt specifications are provided in Appendix~\ref{app:llm_prompts}.

Across 15 model–prompt configurations, posts in the main benchmark were annotated five times using independently constructed random batches and a temperature of 0.1. Repeated annotations were retained to represent variation in support for each value attribution. Configurations were compared with expert annotations. For categorical evaluation model annotations were aggregated by majority vote and evaluated using precision, recall, and F1, together with over- and under-attribution rates (false positives among expert negatives and false negatives among expert positives, respectively). The number of positive LLM annotations (0–5) was retained for graded evaluation against expert vote counts. These criteria were used to select the final model-prompt configuration. Based on these evaluations, Gemini-2.5-pro with the English calibrated prompt was selected for subsequent annotation.
We additionally assessed the stability of the selected configuration by repeating it in two independent sets of five annotation runs and comparing the resulting annotations using Cohen’s $\kappa$ and mean variation in vote counts.

\subsubsection{Error profiling and verification design.} After selecting the final Gemini configuration, but before annotating the training corpus, we used the two expert benchmarks to identify recurrent errors requiring additional human verification. We examined which values were systematically over- or under-attributed, and for these targeted values we identified value co-expression contexts in which these misattributions occurred, and tested whether misattributions were concentrated among lower-confidence LLM judgments. These patterns were used to define targeted verification rules.

\subsubsection{Large-scale annotation and expert verification.} The selected Gemini configuration was applied to 24,000 candidate posts, with each post annotated five times. Annotation labels were converted into soft labels as the proportion of positive LLM votes. The verification rules identified 5,225 potentially problematic annotations. Due to annotation-cost constraints, we randomly sampled 2,000 cases for expert verification proportionally to the size of each flagged subgroup. Each reviewed case was assessed by one expert for the target value and assigned a label of 0, .6, or 1.0 with .6 representing partial evidence of value expression. Expert review resulted in 962 corrections of target-value annotations. 

\subsubsection{Encoder transfer.} The resulting annotated data were divided into 20,000 training and 4,000 test sets. We fine-tuned the last two layers of XLM-RoBERTa-large using mean pooling and a linear multi-label classification head with binary cross-entropy loss (see Appendix~\ref{app:encoder_configuration} for details). We evaluated predictive quality on test set and against the main expert benchmark using Precision–Recall AUC. We also additionally assessed whether encoder scores preserved the confidence of expert judgments by correlating them with expert vote counts. 

\subsection{Corpus-level measures}
We used predictions from the final Gemini-based encoder to characterize value expression in the full candidate corpus. We first calculated post-weighted mean scores for each value. We additionally computed a user-weighted profile by first averaging value-expression scores across all posts of each user and then averaging these user-level means across users, so that each user contributed equally. Hashed user identifiers were additionally matched against the external Botnadzor.org database, which specializes in identifying bot accounts on VK\footnote{URL: https://botnadzor.org. Botnadzor user IDs were hashed using the same procedure applied to VK user identifiers during preprocessing}. We then recomputed the aggregate value-expression profile after excluding matched accounts. To characterize the structure of value co-expression, we calculated pairwise Spearman correlations between value-expression scores at the post level. To examine the structural organization of value co-expression, we derived distances between values from the correlation matrix and applied two-dimensional multidimensional scaling (MDS) with Procrustes rotation toward the theoretical target coordinates\citep{schwartz1992universals}.

\section{Results}

\subsection{Value measurement validation}
As shown in our previous validation study \citep{milkova2026measuring}, LLM annotations had model-specific deviations from expert judgments. For example, under the baseline prompt with short value definitions GPT-4 relatively over-attributed Self-direction and Stimulation, whereas GPT-5.2 and Gemini-2.5 more often attributed Security and Tradition. All three models tended to under-attribute Hedonism and Conformity. Overall, GPT-4 showed lower alignment with expert judgments than GPT-5.2 and Gemini-2.5. Prompt calibration reduced over-attribution patterns and improved agreement with expert judgments. For example, a post expressing willingness to help and support a close person (\emph{“I promise I’ll be there for you — just let me know if you need me”}) was annotated by experts as Benevolence. Under the Baseline prompt, GPT-4 additionally attributed Self-direction, while GPT-5.2 and Gemini attributed Security. After calibration, all models converged on Benevolence only, eliminating these spurious co-attributions. However, in general models respond differently to prompt refinements. GPT-5.2 becomes more conservative (higher precision but lower recall) and Gemini-2.5 maintains a more balanced precision–recall tradeoff, resulted with precision of 0.72, recall of 0.67 and overall F1 of .70. Model--prompt behavior under prompt calibration is shown in
Appendix~\ref{app:model_prompt}.

Importantly, repeated LLM judgments also tracked variation in expert support. On the main benchmark, the mean per-value-specific Spearman correlation between the number of positive Gemini annotations and expert votes was .60. On the second benchmark, Gemini achieved an overall F1 of .64 and a mean value-specific correlation of .64. The relative difficulty of value detection was broadly preserved across the two benchmarks, with Benevolence, Tradition, and Achievement showing stronger agreement and Conformity the weakest. Per-value results are reported in Appendix~\ref{app:final_llm}

Repeated annotations with the final configuration were stable across independently constructed batches (5×5 repeated runs). The final configuration demonstrates high inter-run consistency (Cohen’s $\kappa=.88$) with low average variation in vote counts (mean delta votes = 0.19).

The measurement signal was also preserved after transfer to the encoder used for corpus-level analysis. On the main benchmark, XLM-RoBERTa achieved a macro PR-AUC of .60 against expert annotations, compared with .55 for Gemini on the same dataset. Encoder scores were positively correlated with the number of positive expert votes for every value ($\rho=.33$–$.81$), indicating that predictions preserved graded variation in expert judgments. On the 4,000-post test set, macro PR-AUC against Gemini annotations was .69. Per-value metrics are reported in Appendix~\ref{app:encoder_performance}. We therefore use these scores as the measure of value expression in the analyses that follow.

\subsection{Aggregate value-expression profile}
\begin{figure}[t]
\centering
\includegraphics[width=0.9\columnwidth]{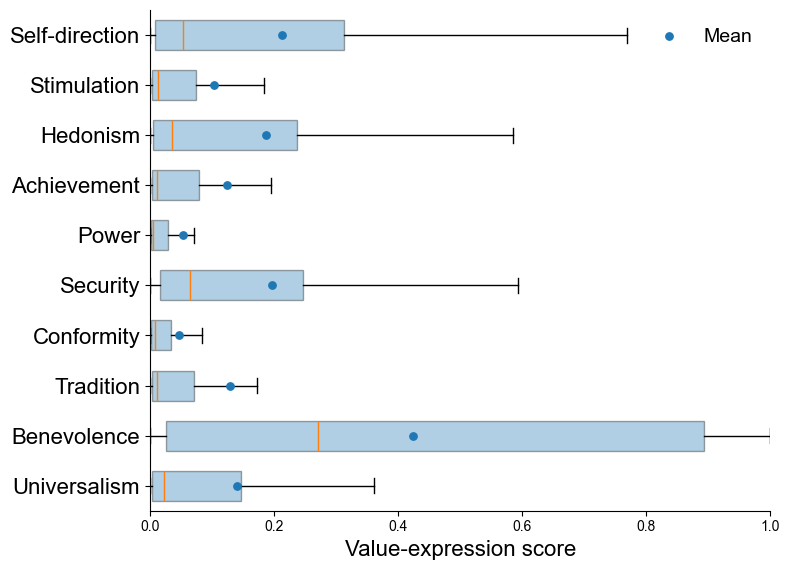} 
\caption{Distribution of post-weighted value-expression scores}
\label{value_scores_boxplot}
\end{figure}

We next characterized the aggregate value profile of public self-expression in Russian social media. In the post-weighted profile, Benevolence received the highest mean score (.424), followed by Self-direction (.214), Security (.197), and Hedonism (.187). Universalism (.140), Tradition (.129), Achievement (.124), and Stimulation (.103) occupied intermediate positions, whereas Power (.054) and Conformity (.047) had the lowest mean scores. 
Value distributions were right-skewed, with mean scores lying above the corresponding medians and long upper tails for all values. Particularly for Stimulation, Achievement, Tradition, and Universalism, relatively low typical scores coexisted with a smaller subset of posts showing much stronger value expression (Figure \ref{value_scores_boxplot}). 

We then examined whether uneven posting activity affected the aggregate profile. Users contributed on average 20.6 value-expressive posts, whereas the median was only 3, and the most active 10\% of users accounted for 76.7\% of all analyzed posts. We therefore compared the post-weighted profile with a user-weighted profile in which each user contributed equally.

\begin{figure}[t]
\centering
\includegraphics[width=0.9\columnwidth]{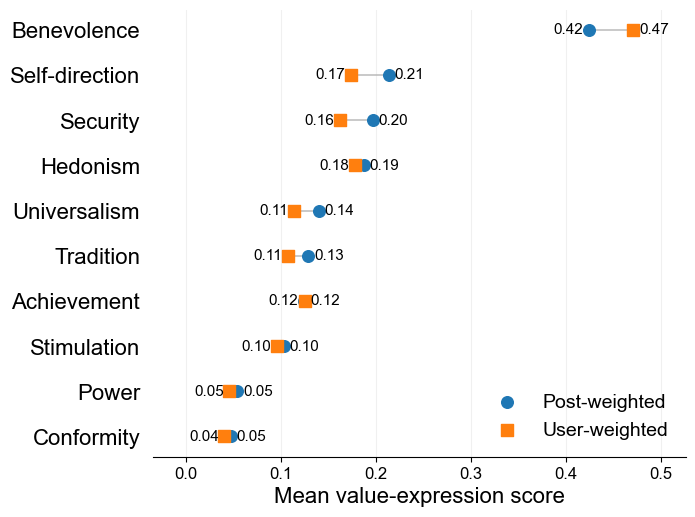} 
\caption{Post-weighted and user-weighted aggregate value-expression profiles.}
\label{value_profile_post_user}
\end{figure}

The overall ordering was similar in the user-weighted profile. Benevolence remained the highest-scoring value, and Stimulation, Conformity and Power – the lowest. However, the ordering of the intermediate values changed. Self-direction, Security, Universalism, and Tradition were relatively more strongly represented in content contributed by highly active users. Benevolence, in contrast, became more prominent when users were weighted equally (See Figure \ref{value_profile_post_user}).  

This value profile is close to the universally \citep{schwartz2001value} and locally  \citep{magun2010life} reported ones: Benevolence often appears the first or the second most important whereas Power values are typically on the bottom. The deviations from the survey-based Russian value profile involve relatively high importance of Hedonism and Self-Direction values and relatively low Universalism. 

A separate analysis of accounts matched to an external database of known bot accounts showed that their exclusion did not materially affect the aggregate profile (see Appendix~\ref{app:bot_sensitivity}).

\subsection{Structure of value co-expression}

\begin{figure}[t]
\centering
\includegraphics[width=0.9\columnwidth]{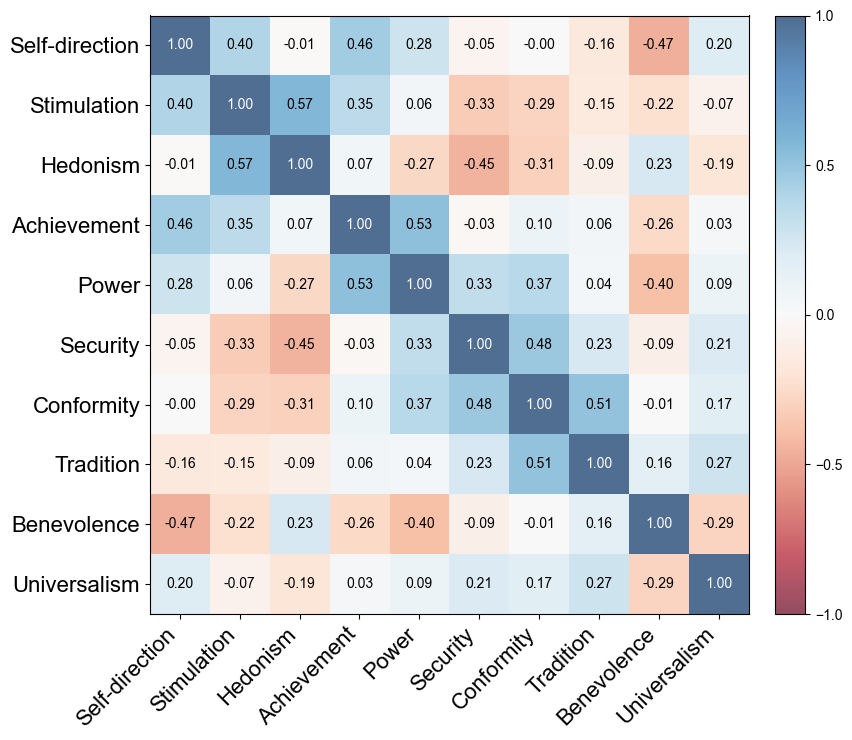} 
\caption{Heatmap of correlations among values.}
\label{value_correlations_spearman}
\end{figure}

We next examined the co-expression of values within individual posts. Most correlations were consistent with Schwartz’s motivational structure (Figure \ref{value_correlations_spearman}). Motivationally compatible values showed positive associations at the post level, including Stimulation–Hedonism ($\rho$=.57), Achievement–Power ($\rho$=.53), Conformity–Tradition ($\rho$=.51), Security–Conformity ($\rho$=.48), and Self-direction–Stimulation ($\rho$=.40), while motivationally incompatible were correlated negatively: Hedonism–Security ($\rho$=-.45), Stimulation–Security ($\rho$=-.33), and Power–Benevolence ($\rho$=-.40).

At the same time, some associations reflected more specific patterns of value expression. Adjacent but theoretically independent Power and Security were positively associated ($\rho$=.33), reflecting right-wing narratives in which political authority and national protection are expressed together. For example: \emph{“Our president is supported by all of Russia! We stand with him and for the victory and protection of our country.”} Universalism was also positively associated with Tradition ($\rho$=.27) and Security ($\rho$=.21). Such combinations appeared in posts where appeals to peace and collective welfare were framed through historical memory and national protection, for example: \emph{“Our fathers once stopped fascism — today we must protect our people and bring peace to every home.”} Another notable deviation was the negative correlation between Benevolence and Universalism ($\rho$=-.29), despite their theoretical proximity within the higher-order Self-Transcendence domain. Thus, concern for close others and concern for people more broadly tended to be expressed in different posts. 

\begin{figure}[t]
\centering
\includegraphics[width=0.9\columnwidth]{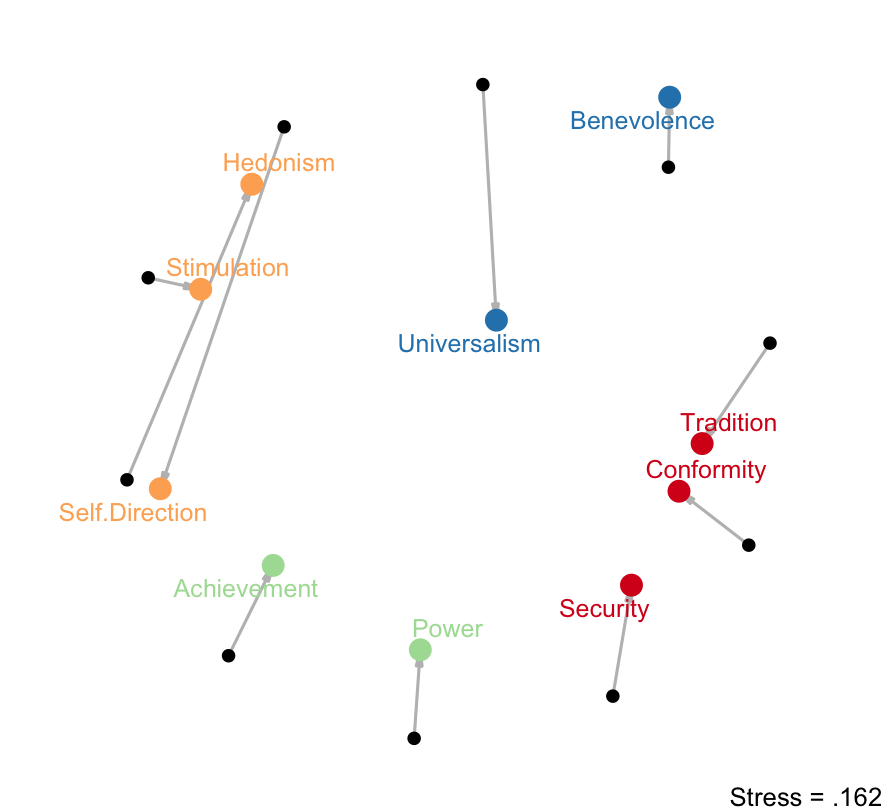} 
\caption{Multidimensional scaling of post-weighted value expressions. Black points represent the ideal theoretical structure in line with Schwartz theory.}
\label{mds_post_level}
\end{figure}

Figure~\ref{mds_post_level} shows the target-rotated MDS of post-weighted value expressions. The congruence coefficients between the resulting coordinates and the ideal Schwartz structure were $\rho_c=.95$ at the post level and $\rho_c=.94$ at the user level, indicating high structural congruence.\footnote{The congruence of survey-based measures with the theoretical structure is high in Russia, as in most other countries. In Russian Social Survey-2021 it was $\rho_c$=.97.}. There are minor deviations: Hedonism and Self-Direction switched places, so that Self-Direction is next to Achievement, while Hedonism is distanced from the other values. This can be explained with the instantiations of Self-Direction which is often expressed in social media via personal achievements. Although this swap is interesting, it occurred within the higher-order value of Openness to Change, so it does not indicate a major deviation from theory. The largest deviation from Schwartz structure is the misallocation of Universalism as an independent entity.

\section{Discussion}

We set out to describe the value structure of spontaneous public self-expression in Russian social media, and to do so with a measurement procedure transparent enough that substantive deviations could be distinguished from artifacts. The validation results indicate that the annotation procedure captures value expression in often ambiguous social media text, with model confidence reflecting the degree of expert agreement in value attribution, and that the resulting value signal can be transferred to an encoder for large-scale measurement.

Two results organize what follows. The circular structure posited by Schwartz's theory is largely recovered from a data source that shares nothing with the questionnaire tradition except the underlying construct. The aggregate profile, by contrast, departs from the survey-based Russian profile in ways that look neither random nor fully systematic, and the deviations are informative in their own right.

Benevolence is the highest-scoring value and Power the lowest, reproducing both the pan-cultural hierarchy \citep{schwartz2001value} and the survey-based Russian profile \citep{magun2010life}. This is not guaranteed by anything in our pipeline. The two measurement modes share no items, no response format, no sampling frame, and no first-person vantage point, so a common ordering at the endpoints is a reasonable indication that both are tracking the same construct. It also supplies the baseline against which the three deviations become readable: they are local displacements within an otherwise familiar hierarchy, not a different hierarchy.

The elevated position of Hedonism in the value profile, compared to the survey-based one, points to the social and identity signaling theories \citep{przepiorka2017signaling, smaldino2022models}. Sharing an enjoyable evening not only expresses the user's own importance of Hedonism, it also sends a signal to the other users. Questionnaires ask how important such motivations are in the abstract; a user timeline gives them something to do. Talking about enjoyment is already a way of pursuing those values, whereas other values may require doing rather than saying. Text-based measures may therefore favor the first kind across the board, a possibility this design cannot rule out. A compositional explanation does not compete. Although VK users are younger than the survey populations, and Openness values are tied to age, the other Openness values, for example, Stimulation scores among the lowest values. 

The implication for the expressed/prompted contrast is that expression and survey scores are not on a common scale even after rank-ordering. Survey instruments make all ten values equally salient and require a response to each, producing a comprehensive and implicitly ipsative profile. Expression data yield an absolute rate of voicing, in which the opportunity structure of the medium is part of the measurement. The two can be compared, but a rank inversion is a candidate sayability effect before it is evidence of different priorities.

Four observations converge on the Universalism value: it is low in the profile; displaced in the structural solution; correlating negatively with its neighbor Benevolence and positively with Tradition and Security; and concentrated in the output of a few highly active users.
 
No single account covers all three. Level and concentration fit together: universalist concerns require an abstract, publicly addressed register that is a minority genre on personal timelines, so Universalism may be weakly expressed rather than weakly held. The structural displacement suggests that universalist vocabulary here is substantially carried by a register framing peace and collective welfare through patriotic and cliched phrasing, rather than the cosmopolitan framing emphasized by the survey items. Indeed, the refined theory's decomposition into concern, nature, and tolerance subtypes gives this a vocabulary, since societal concern is compatible with societal security in a way tolerance is not \citep{schwartz2012refining}.
 
The weighting comparison generalizes this. The most active decile contributes three fourths of posts; Self-direction, Security, Universalism, and Tradition gain under post-weighting,  suggesting that these values are more prominent in an active minority. The value climate an observer encounters is the post-weighted one, and it is measurably more ideological than that of the median user. In contrast, Benevolence is higher under user-weighting, demonstrating a clearly general basic value. 
 
The (modest) discrepancy between post- and user-weighted value profiles warrants general caution. A post is usually a single speech act with a single purpose, so motivationally compatible values may fail to co-occur simply because one post is about one thing. Post-level correlations thus compress positive associations among adjacent values and may manufacture negative ones---the most economical reading of the Benevolence--Universalism negative, where concern for close others and for humanity are competing framings rather than opposing motivations. Since both this exclusivity and measurement error attenuate the configuration, $\rho_c \approx .95$ is better read as a floor than a ceiling.

\section{Limitations and Future Work}

Several limitations concern the scope of the data. First, our dataset was limited to publicly available posts in a state-controlled social network. Given the recent criminalization of certain ideas in Russia, it is plausible that some users preferred making their accounts private or leaving VK. As a result, our sample underrepresents certain types of value expressions or discourses that are less likely to be shared openly.

Second, we analyzed timeline posts but not comments or other interaction data, and the analysis was restricted to textual content. Value expression in conversational settings or in images, memes, and videos may differ from that observed in text-only timeline content. This limits our ability to capture value-related discourse that may occur in different types of data. At the same time, user-initiated posts are more likely to reflect their spontaneous and routine articulations of what matters to users. Nevertheless, future work should extend the framework to conversational and multimodal data.

Third, the present study is limited to Russian-language data and one platform. Applying the approach to other languages or platforms would require new calibration and validation. Our annotation scheme also uses the ten basic values rather than the refined theory of nineteen values (Schwartz et al., 2012), limiting more fine-grained distinctions between value types.

Importantly, the annotations represent theory-informed expert interpretations. They should not be interpreted as universal judgments of how ordinary users perceive value expression.

\bibliography{aaai2026}

\section{Ethics Statement}

This study analyzes only publicly accessible posts from VK profiles that were open at the time of data collection. No private content was accessed, and there was no interaction with social-media users. User identifiers were hashed prior to analysis, and potentially identifying information appearing in post text, including usernames, user mentions, phone numbers, and payment-card numbers, was removed during preprocessing.

Individual consent was not obtained from VK users because the study uses posts from publicly accessible profiles collected through the platform API.

Under the ethical framework applied to the original data-collection study, the project did not meet the criteria requiring formal ethics-committee review. 

We do not release the source corpus publicly, because individual social-media posts may remain searchable and users may face legal or other risks. 

Raw data are stored on an access-controlled server and are accessible only to authorized members of the research team. The server is not directly accessible from the public internet; remote access requires an authenticated VPN connection in addition to password-protected user credentials.

The study focuses on aggregate patterns of value expression, the resulting models should not be interpreted as direct measurements of individuals’ personal values or used for individual-level decision making. The proposed framework is intended for aggregate-level analysis only. Code, prompts, model specifications, and analysis procedures can be shared independently of the source social-media text.

\appendix

\section{Annotation procedures}
\label{app:annotation}
\subsection{Crowd annotation for selection of value-expressive posts}
\label{app:toloka}

Crowd annotation was conducted through the Toloka.ai platform. Participation
was restricted to the top 30\% of workers according to Toloka's internal
rating. Workers were required to correctly label at least 15 of 20
qualification items. Tasks were presented in batches of ten posts with one
control item per batch; workers failing at least two control items or showing
implausibly short response times were excluded. Short ($\leq 35$ words) and
long ($>35$ words) posts were presented separately and compensated at
different rates. A total of 304 Russian-speaking crowd workers participated.
Workers were paid per task, with rates varying by post length. Hourly-equivalent compensation was approximately USD 8 per hour. Total compensation paid to crowd workers was approximately USD 120, excluding platform fees. The original instruction was provided to crowd workers in Russian.
An English translation is reproduced below.
\subsubsection{Instructions provided to crowd workers:}

"This project is for regular users of the VKontakte network!

You are taking part in a scientific study that aims to learn how to
automatically recognize if any expression of life values is contained in
social media posts. Your task is to determine whether there is an expression
of value in the publication that will be shown to you. In other words, you
will be answering the question: \textbf{Does this post reflect the importance
of any idea, phenomenon, or quality?}

\textbf{Yes, it reflects.}
Values are a person's beliefs about the importance of some object, phenomenon,
or quality. Examples of values are: the importance of romantic relationships,
love, fidelity, friendship, and loyalty to loved ones (including love for
children); equality of all people or preservation of nature; stability and
order, following rules, safety, and protecting one's health; independence,
trying new things, and taking risks; achieving goals and receiving deserved
rewards; and simply enjoying life. Other types of values may also appear in
the texts---take into account your own understanding. A text reflecting a
value may not mention the value directly and may also express a negation of a
value; such texts should still be labeled as containing values.

Examples:
\textit{``People are always criticizing. Don't let them break you. Achieving
your goal is the best form of revenge''}---an implicit indication of the
importance of achieving one's goals;
\textit{``The law of meanness is the only law that applies in our country!''}
---an implicit indication of the importance of following rules;
\textit{``This really is one of the best days of our lives! Thank you, my
dear''}---reflecting the value of love;
\textit{``Sometimes you need to treat yourself \#cakes \#cafefroyo
\#eating''}---reflecting the value of pleasures.

\textbf{No, it doesn't reflect.}
The text may not reflect the importance of an object or value. This includes
routine greetings, recipes (culinary or other) without personal evaluation,
and emotional statements that do not indicate the importance or value of
something.

Examples:
\textit{``How touching! I even cried!''} or
\textit{``ay lovely [emoji]''}---the subject of the statement is unclear;
\textit{``Thank you all for the congratulations!''} or
\textit{``new year 2015''}---routine thanks or congratulations.

\textbf{Unclear.}
Texts may be difficult to understand because of unfamiliar names, words,
places, or insufficient context.

\textbf{Spam.}
Texts written for advertising purposes, including on behalf of organizations.
This includes advertisements for goods and services, requests to subscribe,
requests to follow a link, and other promotional posts, including veiled
advertising.

Examples:
\textit{``open sale on direct flight Perm-Min.Vody 226-57-06''}---an obvious
sale announcement;
\textit{``IF YOU WANT TO GO FORWARD, YOU HAVE TO LEARN!!!! And you can do it
with energetic people with positive vibes who are madly in love with their
business and ready to share their knowledge and experience ... <...>
\#depilationorel \#studyingthebest''}---hidden advertising."

\begin{figure}[t]
\centering
\includegraphics[width=0.95\columnwidth]{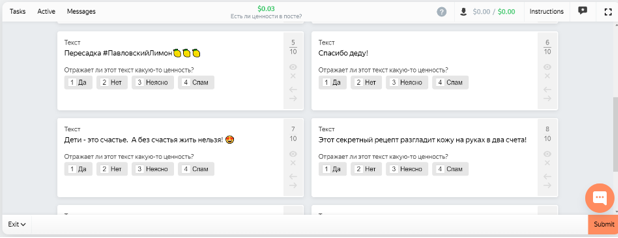}
\caption{Original Toloka annotation interface used for
selection of value-expressive posts. The interface was presented in Russian. The annotation question translates as ``Does this text reflect any value?'', with response options \textit{Yes}, \textit{No}, \textit{Unclear}, and \textit{Spam}.}
\label{toloka_screenshot}
\end{figure}

\subsection{Expert annotation instruction for value classification}
\label{app:expert_instruction}
Expert annotation was conducted in Doccano, an open-source web-based text annotation platform \citep{nakayama2018doccano} released under the MIT License. The original instruction was provided to expert annotators in Russian. An English translation is reproduced below.
\subsubsection{Instructions provided to experts:} "You will be shown posts from the VKontakte social network. Please evaluate
which values are expressed in each post. Evaluate the text itself, rather
than the user's intention.

In this context, values are understood as expressions of the importance of
some object, phenomenon, or quality. A post may express several values or
none of them.

You will answer the following question:

\textbf{Does the post reflect the importance of any objects,
qualities, phenomena, or processes? If yes, which values does it reflect?}

Self-direction: independence of thought and action, with an emphasis
on personal choice, creativity, and broadening one's horizons.

Stimulation: pursuit of novelty, variety, new and intense
experiences, excitement, and risk.

Hedonism: pursuit of pleasure and sensory gratification; the
importance of simply enjoying life.

Achievement: personal success demonstrated through competence in
accordance with social expectations or standards; the importance of achieving
goals and receiving deserved rewards.

Power: social status and prestige, wealth, and control or dominance
over people and resources.

Security: protection, harmony, a strong state, stability of society
and relationships, consistency and continuity; the importance of order,
following rules, safety, and protecting one's health.

Conformity: restraint of actions, inclinations, and impulses that
could violate social norms or laws or harm others. References to laws and
rules were included here. Unlike Tradition, Conformity refers to a more
general tendency to follow rules.

Tradition: respect for and adherence to customs and traditions
prescribed by culture or religion. Unlike Conformity, Tradition focuses on a
specific tradition, such as a religious tradition.

Benevolence: preserving and enhancing the well-being of people with
whom one is in frequent personal contact; the importance of romantic
relationships, love, fidelity, friendship, and loyalty to close others,
including children.

Universalism: understanding, appreciation, tolerance, and protection
of the welfare and equal rights of all people. Unlike Benevolence, the focus is
on people in general rather than close others. Concern for nature is also
included.

A text that reflects a value may not explicitly mention that value. Nevertheless, try to refrain from speculating about the author’s intentions. We are only interested in values that you have no doubt are being expressed. If it’s a very long post, try first to evaluate it as a whole, focusing on the main conclusion—the “moral” of the post.
If a post rejects a certain value, select the “opposite” value. For example, the post “Don’t Be Afraid to Try” implies a rejection of Security, which can be interpreted as the importance of Stimulation. If it is difficult to identify the opposite value, do not select anything.
The texts have already been filtered based on whether they reflect values, so most posts should reflect at least one of the values. However, there are exceptions. In such cases, do not select any of the values.

Examples of posts that reflect values:

\textit{``People are always criticizing. Don't let them break you. Achieving
your goal is the best form of revenge.''} --- Achievement.

\textit{``The law of meanness is the only law that applies in our country!''}
--- an implicit expression of the importance of following rules, classified
as Conformity.

\textit{``This really is one of the best days of our lives! Thank you, my
dear.''} --- Benevolence.

\textit{``Sometimes you need to treat yourself \#cakes \#icecreamcafe
\#eating.''} --- Hedonism.

\paragraph{FAQ}
What should be done with administrative and political posts, such as reports on lawmakers’ activities—for example, that they visited someone, laid flowers, and so on? These posts typically describe the actions of public officials without directly expressing opinions or personal values.

- Try to determine where the emphasis lies. Most often, such posts fall into the “security” category, as they express manifestations of state protection; however, sometimes the emphasis is on equal rights for citizens and care for the poor, in which case the posts may (also) express “universalism.”

What about congratulatory posts?

- Congratulatory posts are the category most heavily laden with values; they directly express what the person offering congratulations considers important. It is not so much a matter of whom these wishes are addressed to, but rather what exactly they contain. In some cases, congratulatory posts are too brief and vague. "

\subsection{LLM annotation instructions for value classification}
\label{app:llm_prompts}

We evaluated five prompt variants: Baseline, Extended, Contextualized,
Bias-calibrated, and Bias-calibrated RU. To avoid repetition, prompt
components that were unchanged across successive variants are reproduced
only once below. All models used the same prompt text within a given
condition.

\paragraph{Baseline prompt.}

You are a professional annotator tasked with categorizing social media posts.
Assign a binary code to each post - whether (code 1) or not (code 0) the post reflects the Values listed below:

Self-direction: Independent thought and action — choosing, creating, and exploring.

Stimulation: Excitement, novelty, and challenge in life.

Hedonism: Pleasure and sensuous gratification for oneself.

Achievement: Personal success through demonstrating competence according to social standards.

Power: Social status and prestige, wealth, control or dominance over people and resources.

Security: Safety, harmony, strong government, and stability of society, relationships, and self.

Conformity: The restraint of actions, inclinations, and impulses that are likely to upset or harm others and violate social expectations or norms.

Tradition: Respect, commitment, and acceptance of the customs and ideas that traditional culture or religion provides.

Benevolence: Preservation and enhancement of the welfare of people with whom one is in frequent personal contact.

Universalism: Understanding, appreciation, tolerance, and protection for the welfare of all people and of nature.

\paragraph{Extended prompt.}

You are a professional annotator of social media posts.
Assign a binary code to each post - whether (code 1) or not (code 0) the post reflects basic human values.
Values are an expression of the importance of an object, phenomenon, or quality. The Values are listed below:

Self-direction: importance of independent thought and action, choosing, creating, and exploring, curious, autonomous.

Stimulation: Excitement, novelty, and challenge in life, daring

Hedonism: Pleasure and sensuous gratification for oneself, enjoying life, self-indulgent

Achievement: Personal success through demonstrating competence according to social standards, ambitious, capable, influential, social recognition

Power: Social status and prestige, wealth, control or dominance over people and resources, authority, social power

Security: Safety, harmony, strong government, and stability of society, relationships, and self, social order, family security, national security, clean, healthy

Conformity: The restraint of actions, inclinations, and impulses that are likely to upset or harm others and violate social expectations or norms, obedient, self-discipline, politeness, responsible

Tradition: Respect, commitment, and acceptance of the customs and ideas that traditional culture or religion provides, humble, devout, moderate, subordination

Benevolence: Preservation and enhancement of the welfare of people with whom one is in frequent personal contact, helpful, honest, forgiving, responsible, loyal, true friendship, mature love

Universalism: Understanding, appreciation, tolerance, and protection for the welfare of all people and of nature, broadminded, social justice, equality, world at peace, protecting the environment.

When assigning values, prioritize conceptual meaning over surface-level lexical cues.
Don't overestimate emotional texts.
Do NOT make assumptions about the author and intentions beyond the text.

Power, achievement, hedonism, stimulation, and self-direction primarily focus on personal interests and characteristics.
Benevolence, universalism, tradition, conformity, and security are primarily concerned with how one relates socially to others and affects their interests.
Power, achievement, tradition, conformity, and security serve to cope with anxiety due to uncertainty in the social and physical world.
Hedonism, stimulation, self-direction, universalism, and benevolence express anxiety-free motivations. Security and conformity aim to avoid or overcome actual or potential danger.
Self-direction and close values motivate intrinsically rewarding social, intellectual, and emotional opportunities.

\paragraph{Contextualized prompt.}

This prompt is identical to the Extended prompt, with the only difference being at the beginning: You are a professional annotator of Russian social media VKontakte born in Russia and living in Russia.

\paragraph{Bias-calibrated prompt.}

You are a professional annotator of social media posts. Assign a binary code to each post - whether (code 1) or not (code 0) the post reflects basic human values.Values are an expression of the importance of an object, phenomenon, or quality. The Values are listed below:

Self-direction: Importance of independent thought and action, making autonomous decisions based on one’s own judgment, creating and exploring, being curious.

Stimulation: Excitement, novelty, and variety in life, daring. Appreciation of surprises, adventures, trying new things.

Hedonism: Importance of having good time, pleasure and sensuous gratification, 'spoiling' themselves, food, leisure, sex or other personally enjoyable activities.

Achievement: Personal success through demonstrating competence according to social standards, importance of being ambitious, capable, influential, getting social recognition.

Power: Social status and prestige, wealth, control or dominance over people and resources, authority, social power. Importance of being visibly rich and in charge.

Security: Safety, harmony, strong government, stability of relationships, of self, of society and social order, national security, clean, healthy.

Conformity: The restraint of actions, inclinations, and impulses that are likely to upset or harm others and violate social expectations or norms.

Tradition: Respect, commitment, and acceptance of the customs and ideas that traditional culture or religion provides, humble, devout, moderate, subordination.

Benevolence: Preservation and enhancement of the welfare of close people with whom one is in frequent personal contact (e.g., family, friends), helpful, honest, forgiving, responsible toward close others, loyal, true friendship, mature love.

Universalism: Understanding, appreciation, tolerance, and protection for the welfare of all people in the world including strangers; importance of broadmindedness, social justice, equality, world at peace. Importance of nature and environment protection.

When assigning values, prioritize conceptual meaning over surface-level lexical cues.
Don't overestimate emotional texts.
If the value is expressed subtly or implicitly, assign it when there is reasonable textual evidence, rather than requiring explicit keywords.
Do NOT make assumptions about the author and intentions beyond the text.

Power, achievement, hedonism, stimulation, and self-direction primarily focus on personal interests and characteristics. Benevolence, universalism, tradition, conformity, and security are primarily concerned with how one relates socially to others and affects their interests. Power, achievement, tradition, conformity, and security serve to cope with anxiety due to uncertainty in the social and physical world. Hedonism, stimulation, self-direction, universalism, and benevolence express anxiety-free motivations. Security and conformity aim to avoid or overcome actual or potential danger. Self-direction and close values motivate intrinsically rewarding social, intellectual, and emotional opportunities.

\paragraph{Bias-calibrated RU prompt.}

The Russian-language condition used the Bias-calibrated prompt in Russian.

\section{LLM annotation behavior and validation}
\label{app:model_prompt}
\subsection{Model-prompt behavior}
\begin{figure*}[t]
\centering
\includegraphics[width=\textwidth]{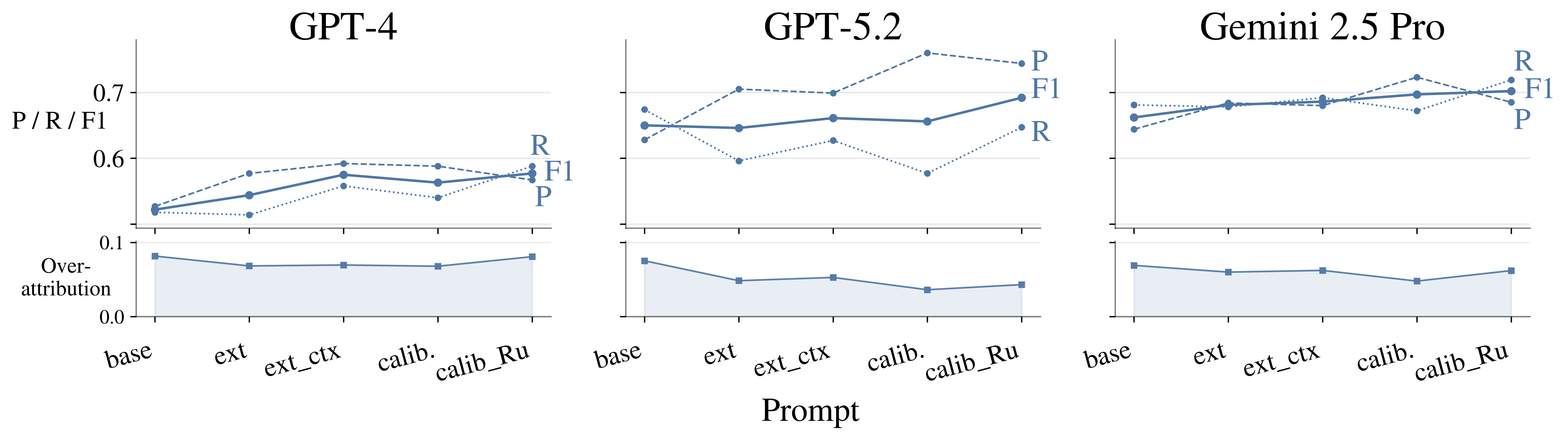}
\caption{Response of GPT-4, GPT-5.2, and Gemini 2.5 Pro to prompt
    refinement on main expert benchmark ($N=1{,}000$). Upper panels show Precision,
    Recall, and F1; lower panels show over-attribution rates.
    \textit{base}: Baseline; \textit{ext}: Extended;
    \textit{ext\_ctx}: Contextualized; \textit{calib.}: Bias-calibrated;
    \textit{calib\_RU}: Bias-calibrated Russian-language prompt.}
\label{P-R-F1-across_model-prompts_blue}
\end{figure*}

Figure~\ref{P-R-F1-across_model-prompts_blue} shows the responses of
different LLMs to prompt refinement.

\subsection{Performance of the final annotation configuration (Gemini-2.5 Pro, Bias-calibrated prompt)}

\label{app:final_llm}

Table~\ref{tab:final_llm_performance} reports per-value performance of the
selected annotation configuration. Benchmark~2 showed lower
precision and higher over-attribution for several values; this dataset also
had substantially lower expert agreement ($\kappa=.30$ vs.\ $.60$ in
Benchmark~1), suggesting a larger share of borderline cases and/or a more conservative expert annotation style. Despite these differences in categorical performance, model vote counts remained positively correlated with expert vote counts for every value in
both datasets.

\begin{table*}[t]
\centering
\begin{tabular}{lcccccc}
\hline
Value & P & R & F1 & OA & UA & $\rho$ \\
\hline
\multicolumn{7}{l}{\textit{Benchmark 1 }} \\
Self-direction & .53 & .70 & .61 & .08 & .30 & .63 \\
Stimulation    & .48 & .54 & .51 & .04 & .46 & .54 \\
Hedonism       & .77 & .49 & .60 & .04 & .51 & .60 \\
Achievement    & .72 & .70 & .71 & .04 & .30 & .68 \\
Power          & .43 & .71 & .53 & .03 & .29 & .52 \\
Security       & .71 & .50 & .59 & .07 & .50 & .60 \\
Conformity     & .57 & .29 & .39 & .01 & .71 & .40 \\
Tradition      & .55 & .85 & .67 & .05 & .15 & .58 \\
Benevolence    & .93 & .83 & .88 & .07 & .17 & .83 \\
Universalism   & .57 & .81 & .67 & .06 & .19 & .67 \\
\hline
\multicolumn{7}{l}{\textit{Benchmark 2 }} \\
Self-direction & .47 & .83 & .60 & .24 & .17 & .64 \\
Stimulation    & .54 & .56 & .55 & .07 & .44 & .51 \\
Hedonism       & .45 & .86 & .59 & .17 & .14 & .74 \\
Achievement    & .64 & .87 & .74 & .12 & .13 & .71 \\
Power          & .49 & .80 & .61 & .12 & .20 & .58 \\
Security       & .39 & .94 & .55 & .33 & .06 & .68 \\
Conformity     & .56 & .41 & .47 & .04 & .59 & .45 \\
Tradition      & .71 & .92 & .80 & .09 & .08 & .79 \\
Benevolence    & .64 & .94 & .76 & .32 & .07 & .72 \\
Universalism   & .29 & .93 & .44 & .17 & .07 & .54 \\
\hline
\end{tabular}

\caption{Per-value performance and graded alignment of Gemini 2.5 Pro (with the English bias-calibrated prompt) on the two expert benchmarks. P = Precision; R = Recall; OA = over-attribution rate; UA = under-attribution rate; $\rho$ = Spearman correlation between
expert positive-vote counts (0--3) and model positive-vote counts (0--5).
All correlations are significant at $p<.001$.}
\label{tab:final_llm_performance}
\end{table*}

\section{Encoder training and evaluation}
\label{app:encoder_configuration}
\subsection{Encoder training configuration}
Table~\ref{tab:encoder_training} shows the fine-tuning configuration for
XLM-RoBERTa-large \citep{conneau2020unsupervised}, initialized from the
\texttt{FacebookAI/xlm-roberta-large} checkpoint available through
Hugging Face and released under the MIT License.\footnote{\url{https://huggingface.co/FacebookAI/xlm-roberta-large}}
\begin{table}[t]
\centering
\begin{tabular}{ll}
\hline
Parameter & Configuration \\
\hline
Model & XLM-RoBERTa-large \\
Fine-tuned layers & Last two layers \\
Pooling & Mean pooling \\
Classification head & Linear multi-label head \\
Loss & Binary cross-entropy (BCE) \\
Batch size & 16 \\
Maximum sequence length & 512 \\
Optimizer & AdamW \\
Learning rate & $2\times10^{-5}$ \\
Early stopping & Validation loss and PR-AUC \\
Maximum epochs & 6 \\
GPU & NVIDIA GeForce RTX 2080 Ti \\
CPU & 48 cores \\
RAM & 128 GB \\
\hline
\end{tabular}

\caption{Training configuration for the XLM-RoBERTa-large encoder.}
\label{tab:encoder_training}
\end{table}

\subsection{Encoder performance}
\label{app:encoder_performance}
Table~\ref{tab:encoder_performance} reports Precision--Recall AUC
(Average Precision) for the XLM-RoBERTa-large encoder. On the main expert benchmark, encoder predictions are evaluated against both expert and Gemini
annotations. Performance on the held-out test set is evaluated against
Gemini annotations used as training targets.

\begin{table}[t]
\centering
\begin{tabular}{lccc}
\hline
 & \multicolumn{2}{c}{Expert benchmark} & Test \\
Value & Experts & Gemini & Gemini \\
\hline
Self-direction & .59 & .74 & .75 \\
Stimulation    & .44 & .65 & .63 \\
Hedonism       & .68 & .67 & .73 \\
Achievement    & .73 & .81 & .79 \\
Power          & .47 & .52 & .47 \\
Security       & .72 & .72 & .69 \\
Conformity     & .23 & .36 & .43 \\
Tradition      & .67 & .80 & .76 \\
Benevolence    & .94 & .94 & .93 \\
Universalism   & .56 & .71 & .70 \\
\hline
Macro average  & .60 & .69 & .69 \\
\hline
\end{tabular}

\caption{Precision--Recall AUC (Average Precision) of the
XLM-RoBERTa-large encoder on the expert benchmark ($N=1{,}000$) and held-out test data ($N=4{,}000$). Columns indicate the reference annotations used for evaluation.}
\label{tab:encoder_performance}
\end{table}
\section{Analysis of bot accounts from Botnadzor database}
\label{app:bot_sensitivity}
Matching user IDs against the Botnadzor database identified 68 accounts in the initial unfiltered corpus, which contributed 17,738 posts. After preprocessing and candidate-post selection, 44 matched accounts remained, contributing 4,144 posts, or 0.37\% of the analyzed corpus. These accounts were substantially more active than other users, contributing on average 94.2 candidate posts per user (median = 15), compared with 20.6 posts on average (median = 3) among other users.

Posts contributed by matched accounts showed a somewhat different
value-expression profile from those contributed by other accounts. Using
post-level mean scores, the largest positive differences were observed for Security (.289 vs. .196), Benevolence (.477 vs. .424), Tradition (.173 vs. .129), and Achievement (.158 vs. .124). Self-direction was less strongly represented among matched accounts (.185 vs. .214), while differences for the remaining values were smaller.

Despite their greater posting activity and different value-expression profile, the matched accounts had negligible influence on aggregate estimates because of their small overall share. Excluding all matched accounts left the ordering of values unchanged, while the largest absolute change in a mean value-expression score was observed for Security (.00035, approximately 0.18\% in relative terms). Thus, accounts identified in this external database did not materially affect the aggregate results.

The relatively small number of matches should be interpreted in light of
the sampling design. Our corpus was constructed from public individual
user timelines and did not include comments or posts collected from
communities. Accordingly, the Botnadzor match should not be interpreted
as an estimate of the overall prevalence of automated accounts on VK.

\end{document}